\documentclass{svproc}
\usepackage{url}
\usepackage{amsmath,amsfonts,amssymb}
\usepackage{graphicx}
\usepackage{setspace}
\usepackage{tocloft}
\usepackage{multirow}
\usepackage[ruled]{algorithm2e}
\usepackage{algorithmicx}

\begin{document}
\mainmatter              
\title{Automated Data-driven Approach for Gap Filling in the Time Series using Evolutionary Learning}
\titlerunning{Data-driven Approach for Gap Filling in the Time Series}  
%
\author{Mikhail Sarafanov\inst{1} \and Nikolay O. Nikitin\inst{1} \and 
Anna V. Kalyuzhnaya\inst{1}}
\authorrunning{Mikhail Sarafanov et al.} 
%
%
\institute{ITMO University, Saint Petersburg, Russia}

\maketitle              


\begin{abstract}
In the paper, we propose an adaptive data-driven model-based approach for filling the gaps in time series. The approach is based on the automated evolutionary identification of the optimal structure for a composite data-driven model. It allows adapting the model for the effective gap-filling in a specific dataset without the involvement of the data scientist. As a case study, both synthetic and real datasets from different fields (environmental, economic, etc) are used. The experiments confirm that the proposed approach allows achieving the higher quality of the gap restoration and improve the effectiveness of forecasting models.

\keywords{time series forecasting, gap filling, machine learning, AutoML}
\end{abstract}
\section{Introduction}
Time series is a common way to represent real-world time process data. As an example of the widely-known applications, different cases of time series can be noted: weather stations, financial stocks, industrial sensors, etc. Due to failures of the sensors themselves of the connection issues, the time series may have gaps. The presence of gaps can be a significant problem for time series forecasting since the vast majority of forecasting models can not proceed with gaps in training data.

The existing open-source solutions (e.g. SSGP-Toolbox~\cite{sarafanov2020machine}) provide only relatively simple methods for the gap-filling. But simple methods can be very inaccurate if the size of the gaps is large. In the paper, we consider an approach for filling gaps based on composite models \cite{nikitin2020structural}. The composite models consist of several machine learning models and can be generated using automated machine learning methods (AutoML). The use of composite models could potentially decrease the error of the forecast. Since such models with several levels can approximate more complex dependencies in the data than single models do.

The restored time series can be used in time series forecasting tasks in the future. Since the restored parts may differ from the original time series, it becomes hard to approximate relationships between elements in the time series. In this case, the forecast based on incorrectly reconstructed historical time series can be inaccurate~\cite{9283191}. To investigate this problem, research has also been conducted within the paper.

The paper is organized as follows. The problem statement is described in Sec.~\ref{sec_ps}. The existing approaches and methods for gap filling in the time series are analyzed in Sec.~\ref{sec_related}. Sec.~\ref{sec_approach} contains the description of the proposed adaptive approach to the gap filling. The results of the experimental evaluation of the proposed approach for different datasets are described in Sec.~\ref{sec_exp}. The main conclusions are proved in Sec.~\ref{sec_concl}.

\section{Problem Statement}
\label{sec_ps}
The analysis of time series is a difficult task if there are a big amount of gaps. If the number of gaps is too large, then trying to restore them will affect the quality of the data-driven forecasting model that is fitted using this data. In practical engineering tasks, it is not the common practice to use time series with the percent of gaps that exceed a certain threshold \cite{weigend2018time}. 

A gap filling problem can be seen as an interpolation problem. Classic definition of interpolation function says that values of interpolation function $\tilde{f}$ should be equal to the values of original function $f$ on a set of primary points ${{x}_{\left\{ i \right\}}},~i\in \left\{ 1,N \right\}$.  

There is a set of points with coordinates on a time axis ${{x}_{\left\{ G \right\}}}$, $~G\in \left\{ t,t+n \right\}$, where $n$ is the number of elements in the gap. In these points we would like to restore values of the time series using information about behavior of the time series before the gap ${{x}_{\left\{ G \right\}}}$ on the set of points ${{x}_{\left\{ H \right\}}},~H\in \left\{ t-{{w}_{1}},t-1 \right\}$ and after – on the set of points ${{x}_{\left\{ N \right\}}},~N\in \left\{ t+n+1,t+n+{{w}_{2}} \right\}$. Here ${{w}_{1}}$ and ${{w}_{2}}$ are the time lags that describe time intervals that contain significant information about behavior of the time series on the interval ${{x}_{\left\{ G \right\}}}$.  

\begin{figure*}[h]
  \centering
  \includegraphics[width=\linewidth]{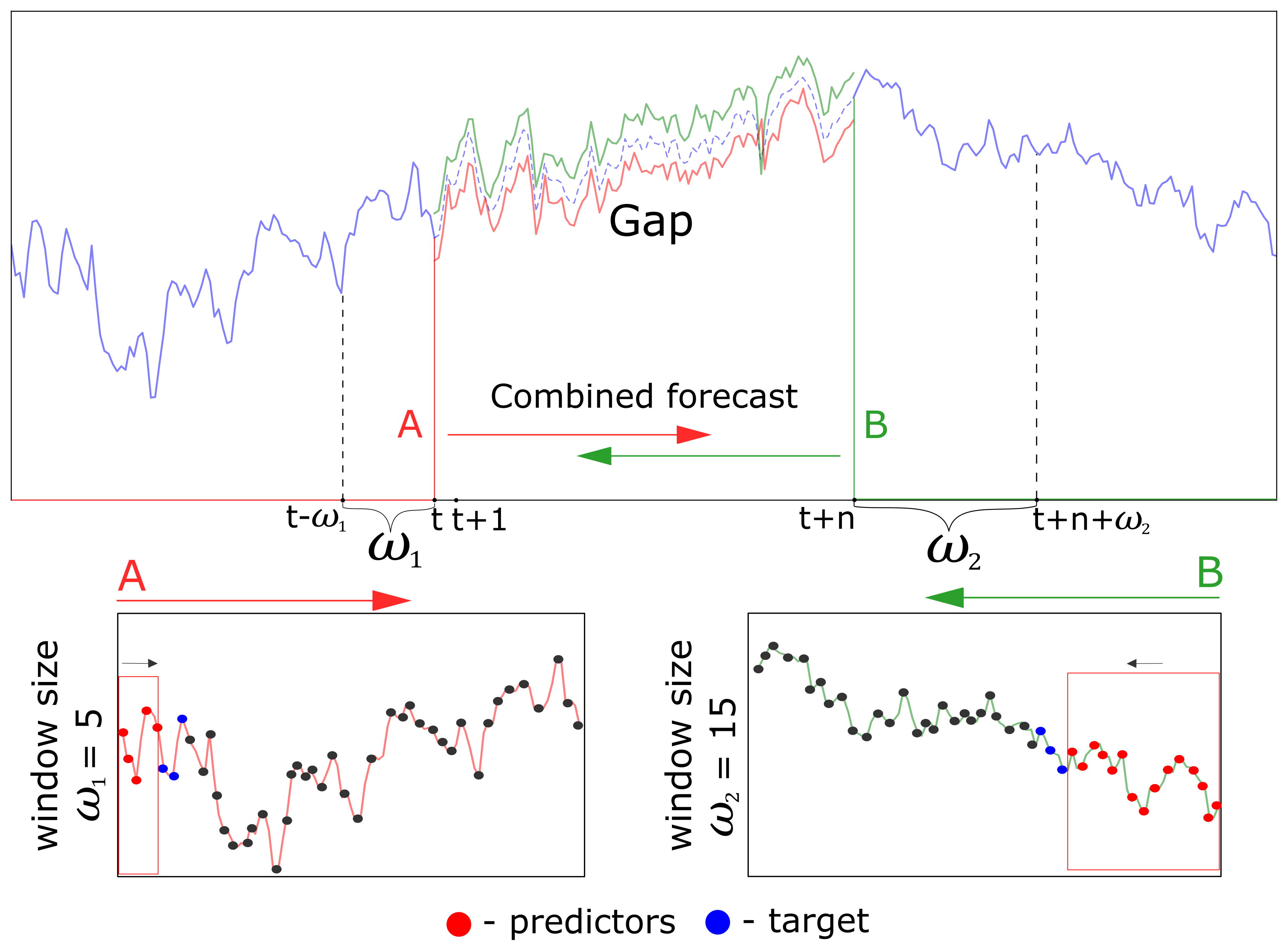}
  \caption{ \label{fig:concept}
    Scheme of the gap-filling approach based at the a bi-directional forecasting. A (red curve) - prediction based on the source part of the time series, B (green curve) - prediction based on the inverted part of the time series. $t$ - time step-index, $w$ - size of the historical time window for forecasting, $n$ - forecast length (less of equal to the gap size).}
\end{figure*}

In the current paper, we suggest an automated data-driven approach that allows working with non-stationary time-series and non-linear predictive models for better gap filling. The main idea of this approach is shown in Fig.~\ref{fig:concept}. In this paper, we aim to study the both efficiency of the AutoML-based models for time series restoration and their impact on the error of the forecasting models that are build using this data.

\section{Related Work}
\label{sec_related}

For time series forecasting tasks there are existing models, for example, convolutional and recurrent neural networks, which can eliminate gaps in data during training~\cite{che2018recurrent}. However, such models are difficult to set up and effectively train. When building neural network models, the amount of data for training is also an important factor - for short time series, it is hard to train complex deep neural networks. Therefore, there are implemented methods that fill in the gaps before applying predictive models.

There are both simple methods, such as linear interpolation and moving average, and more complex ones, for example, spline interpolation and Kalman filters~\cite{lepot2017interpolation}. The disadvantages of simple methods are their inaccuracy for long-term omissions. Partly with this problem can cope more complex algorithms include Radial Basis Functions, Moving Least Squares, Adaptive Inverse Distance Weighted, which are better recovering long skips~\cite{ding2020comparison}. Especially difficult cases can be considered attempts to restore data in time series, where the percentage of gaps exceeds 30\%.

The problem of filling in gaps can also be solved using time series forecasting algorithms. A sequential time-series forecast can be used to fill in the gaps~\cite{7819934}. For such a classical approach autoregressive models, such as AR and ARIMA, can be used. On the other hand, this approach does not take into account the specifics of the gap-filling task. Therefore, a potentially more accurate modification of this approach may be an algorithm that uses both the pre-history and past-history parts of the skip. This approach can be called the forward and backward imputation method~\cite{10.1007/978-3-319-11656-3_9}. For more efficient time series forecasting, evolutionary algorithms can be used as auxiliary tools for selecting important hyperparameters~\cite{lukoseviciute2010evolutionary}. Therefore, in the paper, we relied on evolutionary computing to form an effective algorithm for forward and backward imputation.

\section{Evolutionary design of model-based gap filling approach}
\label{sec_approach}

In this paper combine the existing approaches and automate forecasting-based gap-filling application to increase the quality of the gap-filling.

\subsection{Model-based gap filling with ML and AutoML}

The main idea of the proposed approach is the following. The problem of gap filling in the time series can be reduced to the well-studied problem of time series forecasting. That is, use only the data before the gap (pre-history) to configure the model, and then apply it to get a sequence of values of the same size as the length of the gap. To build the model, AutoML approaches can be used.

However, in this case, the specifics of the task are not taken into account and the part of the time series after the gap (post-history) is not used. To resolve this issue, we have used the bi-directional approach to restore values (the pseudo-code of the underlying algorithm provided in Alg.~\ref{alg:bidir}).

\begin{algorithm}[ht!]
\SetAlgoLined
\KwData{time\_series\_with\_gaps;\\
$w$ $\leftarrow$ moving window size;\\
}
\KwResult{time\_series\_without\_gaps}

 gaps $\leftarrow$ all gap-induced segments in time\_series\_with\_gaps\\
 correct\_data $\leftarrow$ gap-free part of time\_series\_with\_gaps\\
  
  time\_series\_without\_gaps = time\_series\_with\_gaps \\
 \For {gap in gaps} {
  gap\_id $\leftarrow$ index of gap segment start\\
  pre\_window $\leftarrow$ subsec(gap, -$w$) \\
  post\_window $\leftarrow$ subsec(gap, $w$) \\
  models $\leftarrow$ generate\_population\\
    \For {generation in amount\_of\_generations}{
    models $\leftarrow$ apply\_mutation(models) \\
    models $\leftarrow$ apply\_crossover(models) \\
    best\_models $\leftarrow$ select\_fittest(models, pre\_window, post\_window) \\
    models $\leftarrow$ apply\_reproduction(best\_models) 
    }
    gap\_filling\_model $\leftarrow$ select\_best\_model(correct\_data) \\
    
  forward\_prediction $\leftarrow$ gap\_filling\_model.predict(pre\_window,) \\
  backward\_prediction $\leftarrow$ gap\_filling\_model.predict(post\_window)
  
  prediction $\leftarrow$ ensemble\_model(forward\_prediction, backward\_prediction)
  
  time\_series\_without\_gaps[gap\_id] = prediction
 }
 \caption{ \label{alg:bidir} Pseudocode of the algorithm for the bi-directional gap filling based at composite model, obtained using evolutionary optimisation. The full description of evolutionary part is provided in Fig.~\ref{fig:fedot}}
\end{algorithm}

In this case, all possible information available in the time series is used to fill in the gaps. This approach is potentially more accurate but requires more computing resources.

\subsection{Composite modelling}

The data-driven bi-directional model for gap-filling can have a complex structure. Models can be combined into ensembles or stacked into multi-level pipelines, where predictions from one level of models can be predictors for the next level. These structures are naming composite models and can be effectively generated using evolutionary structural learning \cite{nikitin2020structural}.
 
The proposed self-adapting gap-filling algorithm is implemented on a basis of the open-source automated modeling framework FEDOT\footnote{https://github.com/nccr-itmo/FEDOT}. It allows building the data-driven and hybrid models consist of several atomic blocks~\cite{nikitin2021automated}. Regression models on lagged features (e.g. lasso regression or K-nearest neighbors) can be used as such blocks.
 
\begin{figure*}[h]
  \centering
  \includegraphics[width=\linewidth]{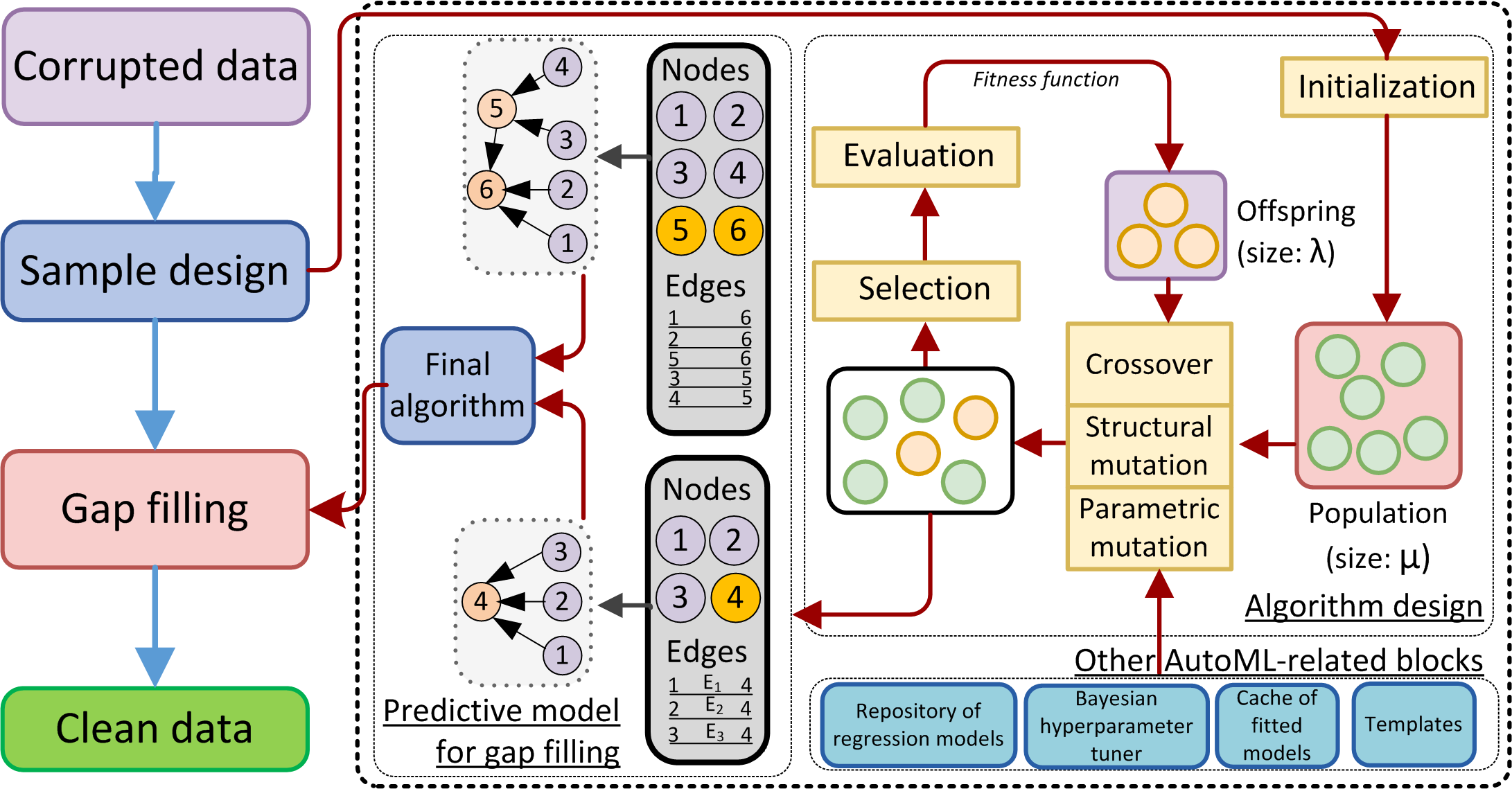}
  \caption{ \label{fig:fedot}
    The pipeline of the evolutionary design of gap filling algorithm for specific problem using AutoML-based techniques. The combination of the modelling blocks and data preprocessing blocks allow obtaining the composite predictive model that is used in the algorithm.}
\end{figure*}

The evolutionary model design is implemented on a basis of the custom graph-based evolutionary approach. The common pipeline of adaptive evolutionary-based gap filling is presented in Fig.~\ref{fig:fedot}.

The Fig.~\ref{fig:fedot} shows that for each time series with gaps (corrupted time series), an algorithm for automatic model design is started to restore the values. During this algorithm's execution, the population with models is initialized. And then the crossover and mutation operators are applied to it. As a result of the structure of the composite model search and tuning the hyperparameters, the final model (Gap filling algorithm) is ready for use.

We implement both basic methods of gap-filling and the proposed model-based approach as parts of the framework. These implementations and different examples of its applications are available as open-source code.

\section{Experimental Study}
\label{sec_exp}

To validate the proposed approach, we run a set of experiments using several datasets with different properties. We used an artificially generated (synthetic) time series, a time series of sea surface height with hourly and daily discreteness, time series with air temperature and economic time series.

\subsection{Datasets}

Five time series of different nature were prepared: a synthetic time series (1) sea level (2) time series obtained from the reanalysis grid of satellite altimetry, a sea level (3) time series obtained from simulating the sea surface height in the Arctic Ocean, time series of the air temperature (4), economic time series (5). 

\begin{figure}[h]
\centering
\includegraphics[width=\linewidth]{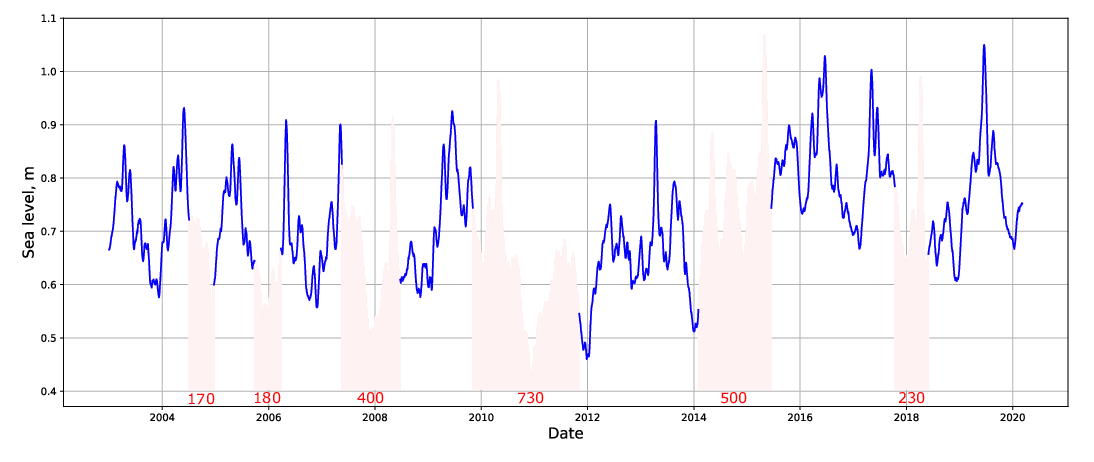}
\caption
{ \label{fig:gaps}
Examples of generated gaps. The total length of gaps is 2210 elements. Red labels indicate the number of time series elements in the gap.}
\end{figure}

However, the synthetic data with desired properties can be obtained with an equation-based model that may be approximately restored from the real-date using algebraic terms approach \cite{merezhnikov2020closed} (if necessary). The length of each obtained time series is 6276 elements. Synthetic gaps were generated with a total size of 30\% of the length of the time series (an example of time series with synthetic gaps can be seen in Fig.~\ref{fig:gaps}). Also, the long gap (1500 steps) was generated in the central part of the time series to assess the errors of data recovering for long gaps. 

For the reconstructed sections of the series, the predicted values were compared with the actual using Mean Absolute Percentage Error (MAPE) and symmetric MAPE (SMAPE). After gap-filling procedure the reconstructed time series were used to train predictive models, and then a forecast was given. For validation, MAPE measures on the sequence of 20\% of the length of the time series were used. The more the error of the forecast increases, the worse the reconstructed time series is suitable for model training.

\subsection{Experimental setup}

There are different algorithms that were applied to the datasets. The list of algorithms, involved in the experimental studies is the following:
\begin{itemize}
\item Linear interpolation - as baseline;
\item Local approximation by polynomial functions (Savitzky–Golay filters), Batch approximation by polynomial functions - as simple approximation-based approaches. The polynomial batch approximation differs from the local one in that the batch approximation use one polynomial function for the entire interval with gaps, no matter how many elements are missed in it. And with local approximation, a polynomial is constructed for each gap element (one polynomial function per one element);
\item Kalman filter, moving average, spline interpolation - as methods from widely-used library "imputeTS"~\cite{RJ-2017-009};
\item Non-linear time series forecasting model, which iteratively predict missing parts in time series. The pipeline was identified manually and comprise of a lagged transformation and a decision tree model;
\item Composite model identified by AutoML - as an approach based on the framework FEDOT. 
\end{itemize}

To make the research more valuable and reproducible, we implemented the evolutionary approach described in the paper as a part of the FEDOT AutoML framework functionality. 

\subsection{Results of the experiments}

The average values for each time series are shown for several cases shown in Table~\ref{tab:results}. The table also contains information about forecasting errors when the restored time series was used.

\begin{table}
\centering
\caption{The results for estimating the error of gap-filling algorithms using MAPE measure for time series with different parameters. The results of the error of the predictive model using the reconstructed time series are shown}
\label{tab:results}
\begin{tabular}{|c|c|c|c|c|c|c|} 
\hline
\multirow{2}{*}{Algorithm}                                                    & \multicolumn{5}{c|}{MAPE}                                                                                                                                                                                                                             & \multirow{2}{*}{\begin{tabular}[c]{@{}c@{}}MAPE of \\forecast on \\restored series\end{tabular}}  \\ 
\cline{2-6}
                                                                              & Synthetic     & \begin{tabular}[c]{@{}c@{}}Sea level,\\hourly\end{tabular} & \begin{tabular}[c]{@{}c@{}}Sea level,\\daily\end{tabular} & \begin{tabular}[c]{@{}c@{}}Tempera-\\ture\end{tabular} & \begin{tabular}[c]{@{}c@{}}Eco-\\nomic\end{tabular} &                                                                                                   \\ 
\hline
\begin{tabular}[c]{@{}c@{}}Linear\\interpolation\end{tabular}                 & 16.0          & 20.3                                                       & 14.4                                                      & 3.5                                                    & 193.1                                               & 11.6                                                                                              \\ 
\hline
\begin{tabular}[c]{@{}c@{}}Local\\approximation by~\\polynomials\end{tabular} & 74.4          & 156.3                                                      & 181.6                                                     & 9.5                                                    & 141.8                                               & 13.6                                                                                              \\ 
\hline
\begin{tabular}[c]{@{}c@{}}Batch\\approximation by \\polynomials\end{tabular} & 28.5          & 50.6                                                       & 57.2                                                      & 3.6                                                    & 174.1                                               & 12.4                                                                                              \\ 
\hline
Kalman filter                                                                 & 41.4          & 105.5                                                      & 151.4                                                     & 22.2                                                   & 247.9                                               & 17.3                                                                                              \\ 
\hline
Moving average                                                                & 18.4          & 14.1                                                       & 25.1                                                      & 4.0                                                    & 205.9                                               & 11.8                                                                                              \\ 
\hline
\begin{tabular}[c]{@{}c@{}}Spline\\interpolation\end{tabular}                 & 223.0         & 394.6                                                      & 312.6                                                     & 31.3                                                   & 146.5                                               & 13.6                                                                                              \\ 
\hline
Non linear                                                                    & 13.6          & 17.9                                                       & 15.1                                                      & 3.3                                                    & 136.4                                               & 12.0                                                                                              \\ 
\hline
\begin{tabular}[c]{@{}c@{}}\textbf{Proposed }\\\textbf{approach}\end{tabular} & \textbf{11.9} & \textbf{16.7}                                              & \textbf{14.9}                                             & \textbf{2.7}                                           & \textbf{32.4}                                       & \textbf{11.2}                                                                                     \\
\hline
\end{tabular}
\end{table}

\begin{figure}[h]
\centering
\includegraphics[width=\linewidth]{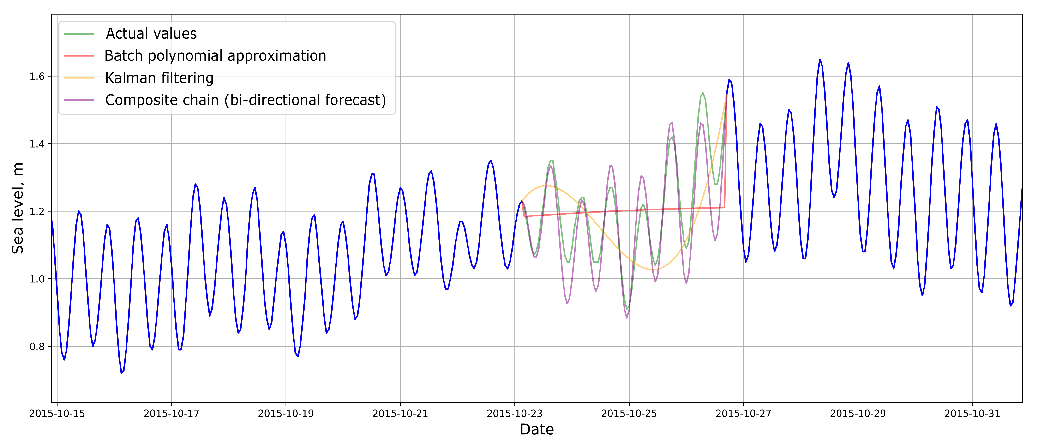}
\caption
{ \label{fig:filled}
Examples of time series restoration by different gap-filling approaches for the sea surface height dataset with hourly discreteness.}
\end{figure}

\begin{figure}[h]
\centering
\includegraphics[width=\linewidth]{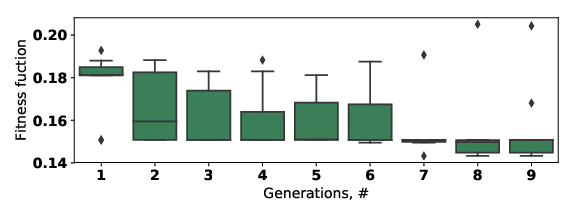}
\caption
{ \label{fig:evo_conv}
The convergence of the RMSE-based fitness function during the evolutionary optimization of the composite model structure for the gap-filling task (sea surface height case).}
\end{figure}

As can be seen, the best results were obtained based on the composite AutoML model (Fig.~\ref{fig:filled}). The composite model has well reconstructed the phases of fluctuation of the height of the sea surface(the convergence of the structural learning of the composite model is demonstrated in Fig.~\ref{fig:evo_conv}). According to SMAPE measure, the following list ranked from best to the worst was obtained: proposed approach (16.4\%), linear interpolation (48.4\%), moving average (55.4\%), batch approximation by polynomials (64.0\%), local approximation by polynomials (110.8\%), Kalman filter (115.9\%), spline interpolation (231.6\%). 

The composite model has well reconstructed the phases of fluctuation of the height of the sea surface (the convergence of the structural learning of the composite model is demonstrated in Fig.~\ref{fig:evo_conv}).

\section{Conclusions}
\label{sec_concl}

An approach to fill gaps in time series, using automatic machine learning methods was presented and validated. It was found that the most effective approach was bi-directional forecasting using the evolutionary algorithm with automatic identification of the model. The obtained AutoML-based solution got averaged MAPE of 15.7\% and SMAPE of 16.4\% for the gap-filling task,  while the competitive algorithms could not have error less than 49.5\% MAPE (48.4\% SMAPE). For a synthetic time series with a breakpoint (when the periodicity of one of the component components changed), the bi-directional approach also can be considered as the better solution.

The reconstructed time series were used to fit forecasting models and predict new values of the time series. The fitting forecasting model on time series reconstructed by the proposed approach distorted the forecasts less than using the others series. It confirms that the proposed data-driven approach in conjunction with AutoML techniques allows efficiently recovering gaps in time series.

It is planned to explore the possibilities of using the AutoML approach to restore multivariate time series. It is also planned to improve the performance of the proposed approach.

\section*{Code and data availability}

All implemented approaches are available in repository \url{https://github.com/nccr-itmo/FEDOT} as a part of the open-source FEDOT framework and can be used for practical purposes. Data and scripts used to conduct the experiments in the paper are available in the additional repository\footnote{\url{https://github.com/ITMO-NSS-team/FEDOT-benchmarks/tree/master/experiments/gap_filling}}.

\section*{Acknowledgements}
This research is financially supported by The Russian Science Foundation, Agreement \#17-71-30029 with cofinancing of Bank Saint Petersburg.

\bibliographystyle{splncs03}
\bibliography{main}

\end{document}